\documentclass{article}

     \usepackage[final]{neurips}

\usepackage[utf8]{inputenc} 
\usepackage[T1]{fontenc}    
\usepackage{hyperref}       
\usepackage{url}            
\usepackage{booktabs}       
\usepackage{amsfonts}       
\usepackage{nicefrac}       
\usepackage{microtype}      
\usepackage{graphicx}
\usepackage{subcaption}

\title{The effects of gender bias in word embeddings on depression prediction}

\author{
  Gizem Sogancioglu\\
  Utrecht University\\
  Utrecht, Netherlands \\
  \texttt{g.sogancioglu@uu.nl} \\
  \And
  Heysem Kaya \\
  Utrecht University \\
  Utrecht, Netherlands \\
  \texttt{h.kaya@uu.nl} \\
}

\begin{document}

\maketitle

\begin{abstract}
Word embeddings are extensively used in various NLP problems as a state-of-the-art semantic feature vector representation. Despite their success on various tasks and domains, they might exhibit an undesired bias for stereotypical categories due to statistical and societal biases that exist in the dataset they are trained on. In this study, we analyze the gender bias in four different pre-trained word embeddings specifically for the depression category in the mental disorder domain. We use contextual and non-contextual embeddings that are trained on domain-independent as well as clinical domain-specific data. We observe that embeddings carry bias for depression towards different gender groups depending on the type of embeddings. Moreover, we demonstrate that these undesired correlations are transferred to the downstream task for depression phenotype recognition. We find that data augmentation by simply swapping gender words mitigates the bias significantly in the downstream task.

\end{abstract}

\section{Introduction}
Gender differences and biases towards minorities in healthcare are heavily studied by researchers. The source of bias in healthcare can be due to multiple reasons such as gender differences in clinical trials and research~\cite{holdcroft2007gender}, professionals' unaware unfair treatment towards minorities~\cite{deangelis2019does} or diagnosis based on symptoms of majority groups~\cite{arslanian2006symptoms}. Mental disorders are one of the healthcare categories that are heavily affected by societal and cultural norms. While many studies report gender inequalities~\cite{doering2011literature} in the diagnosis of depression/anxiety, researchers also found that women take significantly more prescribed psychotropic drugs compared to men~\cite{bacigalupe2021gender}. These societal or statistical biases that exist in the real world and existing training resources are carried by ML models and consequently, this causes an unfair treatment of sensitive groups e.g., based on gender or race.

Bias can be exhibited in multiple parts of the natural language processing (NLP) models; from training data, and pre-trained word embeddings which are the core of state-of-the-art NLP models~\cite{sogancioglu2022gender}, resources, and algorithms themselves~\cite{chang2019bias}. In addition, the final model's predictions can even amplify the biases present in the part of the pipeline~\cite{mehrabi2021survey}. After the striking findings about gender bias in word embeddings for stereotypical occupations (e.g. female vectors are closer to nurse, while the male vectors are to doctor) by the study~\cite{bolukbasi2016man}, the research on the fairness of embeddings has accelerated. Many studies focused on various sub-problems of fairness such as quantifying bias in embeddings~\cite{caliskan2017semantics, kurita2019measuring}, fairness analysis in contextual~\cite{kurita2019measuring, basta2019evaluating, zhao2019gender} or non-contextual embeddings~\cite{bolukbasi2016man}, methods for de-biasing embeddings~\cite{kaneko2019gender}, and many more. Similar fairness analyses are also applied to clinical domain-specific embeddings~\cite{zhang2020hurtful, agmon2022gender} and researchers studied bias in downstream clinical tasks such as in-hospital mortality prediction~\cite{zhang2020hurtful} and depression research using social media~\cite{aguirre2021gender}.  

Although there are many machine learning (ML) studies on depression diagnosis in the literature, there has been a little attempt on analyzing these models in terms of fairness. A recent study analyzes the fairness of depression classifiers trained on social media, however, their focus is on the bias that exists in available depression datasets~\cite{aguirre2021gender}. Therefore, our research question is whether word embeddings, which are commonly used in depression diagnosis~\cite{trotzek2018word, mallol2019hierarchical}, are gender biased for the depression category and if so how these biases are translated to the depression-related downstream tasks. 

We summarize our contributions to this study as follows; 
\begin{itemize}
    \item For the first time, we analyze fairness in word embeddings for the depression category. 
    \item We show that gender bias direction changes based on the dataset used to train embeddings. 
    \item We perform a set of experiments repeated with different algorithms to examine the downstream effects of bias in embeddings using the depression phenotype recognition task. 
\end{itemize}

\section{Experimental Validation}
In this section, we first explain the publicly available pre-trained embeddings used in our experiments and measures to quantify bias in embeddings. Then, we give details about the depression phenotype recognition task and the designed experiments to measure fairness in ML models.  
\subsection{Fairness Analysis in Embeddings}
We used four different pre-trained embeddings in our experiments which are summarized below.
\newline \newline
\textbf{W2VecNews}~\cite{mikolov2013distributed} embeddings are trained on a part of the Google News dataset and contain 300-dimensional vectors for 3 million words and phrases. \newline
\textbf{BioWordVec}~\cite{chen2019biosentvec} are FastText~\cite{bojanowski2017enriching} embeddings trained on PubMed corpus~\footnote{available from https://github.com/ncbi-nlp/BioWordVec}. \newline
\textbf{Clinical-BERT}~\citep{alsentzer2019publicly} embeddings were trained on all available clinical notes of the MIMIC-III Clinical Database which consists of the medical notes describing the diagnosis and treatment of 46.520 patients at the Intensive Care Unit~\citep{johnson2016mimic}. \newline
\textbf{W-BERT}~\cite{devlin2018bert} is word-level contextualized BERT embeddings that are trained on Wikipedia and book corpus with masked language modeling objective. We used the `bert-base-cased' model. 
\subsubsection{Direct Bias}
To quantify bias in embeddings, we used the Direct Bias (DB) measure~\cite{bolukbasi2016man}, which is also used in many previous studies for both non-contextual and contextual embeddings~\cite{basta2019evaluating}. \textit{Direct bias} is computed by averaging the cosine similarity scores between the gender vector and the words belonging to the target category.
We used the depression synonym list that was extracted by a recent study~\cite{koleck2021identifying} as our target category list. The list consists of symptom-related words such as `depressed', `anxiety', and `falling asleep'. 

To compute the gender vector, we used gender pairs list\footnote{\url{https://github.com/tolga-b/debiaswe/blob/master/data/definitional_pairs.json}} (e.g her-him, she-he). The difference vectors of the ten gender pairs are fed into the principal component analysis (PCA). The first eigenvector, which explains the majority of variance, represents a gender direction. The average absolute cosine similarity score between each word in the depression synonym list and the gender vector gives a DB score for the depression category. 

The word vectors were extracted by using the aforementioned pre-trained embeddings. However, since contextual embeddings require context to obtain vectors for the given word, we created sentences containing gender or depression words. For gender pairs, we constructed simple sentences by swapping given gender pairs (e.g. he is a man, she is a woman). For depression words, we used a template that does not contain any gender pronouns yet can be used as a simple explanation of terms: "\textit{X is a synonym of depression.}".  

\subsection{Downstream Task: Depression Phenotype Recognition}
Evaluating the effect of bias in embeddings on depression-related problems is a difficult task as the bias can be transferred from multiple parts of the NLP model. To simplify our analysis, we needed a dataset in which gender information is explicitly given by gender pronouns but possibly does not exist in other words.
\paragraph{MIMIC-III-subset} 
We used the subset of MIMIC-III~\cite{johnson2016mimic} clinical notes events, which were annotated by two human experts~\citep{moseley2020phenotype} for 15 clinical patient binary phenotypes. The clinical notes are written by a 3rd person (e.g. nurse or practitioner), thus gender pronouns to refer to the patient are extensively used. Since our focus in this study is depression, we only included the notes that were either labeled with `depression' phenotype or `none' phenotype. `None' label means that no indication or cue was apparent to the annotator. This resulted in a total of 672 labeled clinical notes. For the training set, we used 90 `depression' and 90 `none' labeled notes for each gender group, in order to minimize bias towards a class and gender-group. In the depression phenotype recognition model, we expect the model to behave the same to the same notes with different gender pronouns. To measure the fairness from this angle, the remaining 312 notes were doubled by swapping gender pronouns with the opposite group's pronouns (e.g. he->she, him->her). 

\paragraph{Experimental design} 
To evaluate the effect of bias in embeddings to downstream tasks, we define 4 different experiments;

1. \textit{original}: train a binary classifier on the original training data.
2. \textit{swapped}: train a binary classifier on the training dataset in which gender pronouns in the original data were swapped with other gender group's pronouns. 
3. \textit{neutralized}: neutralize the data by either removing or replacing all gender pronouns with gender-neutral pronouns and train the classifier on this neutralized set. 
4. \textit{augmented}: train a binary classifier on the union of the original and swapped datasets. 

We repeat the same set of experiments with different learners including Support Vector Machine (SVM), Random Forest (RF), and Multilayer Perceptron (MLP) to validate whether the findings are algorithm-specific. Hyperparameter tuning was done by 3-fold cross-validation on the training set for every model. For SVM, we only tuned C parameter (in [0.01, 100] range with exponential steps) and kernel (in \{rbf, sigmoid, linear\}). For RF, we tuned the maximum depth of the tree in the range of [1-50]. And, for MLP, we used 1 hidden layer having 100 neurons with ReLU activation and only tuned alpha parameters in the range of [0.1, 10].  

\section{Experimental Results}
\begin{table*}
\begin{tabular}{lllllllllllll}
\hline
\multicolumn{1}{|l|}{Model} & \multicolumn{1}{|l|}{DB} & \multicolumn{2}{l|}{original} & \multicolumn{2}{l|}{swapped} & \multicolumn{2}{l|}{neutralized} &\multicolumn{2}{l|}{augmented}\\ \hline
          & &  FNRR     & F1  & FNRR   & F1  & FNRR    & F1  & FNRR    & F1  \\
Clinical-BERT   & 0.04$_M$  & 0.81$_F$ &  0.64 & 0.61$_M$ &  0.66 & 0.62$_M$  & 0.64 & 0.97$_F$ &0.63 \\
W-BERT  &  0.02$_F$  & 0.95$_F$ & 0.58 & 0.93$_M$ & 0.58 &  0.76$_F$ & 0.59 & 0.97$_F$ & 0.59 \\ 
BioWordVec   & 0.09$_M$ & 0.90$_M$ & 0.63 & 0.96$_M$ & 0.62 & 0.72$_M$ & 0.61 & 0.97$_M$ & 0.67 \\ 
W2VNews   &  0.04$_F$  & 0.90$_F$ &  0.61 & 0.95$_F$ & 0.61& 0.36$_F$ & 0.60&  0.97$_M$ &0.63 \\ 
\hline
\end{tabular}

\caption{\label{tab_res}Performance measures of trained SVM models alongside bias measures of embeddings. DB: Direct bias score of given word embedding, if embedding is biased towards the female, denoted with subscript $_F$, otherwise with $_M$, F1: macro-averaged F1 score. FNRR: false negative rate ratio. If the female group's FNRR is lower than the male group, this is denoted with $_F$, otherwise with $_M$.}
\end{table*}

Direct Bias scores of each embedding method, alongside performance measures of trained SVM classifiers are given in Table~\ref{tab_res} (for the results of MLP and RF models, see the Section~\ref{sec:supp}). 
While we observe gender bias in all pre-trained embeddings, the magnitude of these biases varies. Although we cannot directly compare the contextual and non-contextual embeddings, we observe that BioWordVec carries a higher bias than domain-independent W2VecNews. At the same time, Clinical-BERT is more biased towards gender groups compared to domain-independent W-BERT. 
Another interesting finding is that the bias direction is opposite between embeddings trained on the clinical dataset (closer to male) and domain-independent set (closer to female). Although depression is more prevalent in females based on medical literature, clinical embeddings trained on PubMed articles or MIMIC-III datasets show bias towards to male group.

Depression phenotype recognition models trained on original data have 10 to 30 mismatch predictions over 312 clinical note pairs with opposite gender pronouns. In other words, the model gives different predictions up to 30 notes when we change only gender pronouns in the test examples. This result motivates us to further analyze the bias in these models. To observe the effect of this bias direction exist in embeddings, we need to analyze the differences in 4 experiments namely \textit{original, swapped, neutralized, and augmented}. Let's assume that none of the pre-trained embeddings are biased. In this case, 1. we expect to see a very similar score between \textit{original} and \textit{swapped} experiments with a change in the bias direction. Moreover, 2. for the \textit{neutralized} experiment, we expect to see a very high false negative rate ratio (FNRR) between gender groups. Because, training data is free of gender information and consequently, ML models will not learn any undesired relations between gender pronouns and depression. Regarding the first point, we observe that the gender bias direction either does not change or the FNRR score changes largely for BioWordVec and W2VecNews embeddings for all the experiments with 3 different learners. Regarding the second point, we observe consistently lower false negative rate scores for the gender group that the embedding of the model is biased towards. These findings support the bias measures in the embeddings. Based on these results, we can say that gender bias in embeddings is transferred to downstream tasks by favoring one group with less false negative rates. Moreover, we expect to see improved FNRR with \textit{augmented} experiment as the model is taught to make no difference based on gender pronouns by using identical notes with swapped gender. Similar to findings reported in~\cite{zhao2018gender}, a simple augmentation approach (shown in the \textit{augmented} experiment) improves the fairness of the models with an above 90\% FNRR. 

On the other hand, based on these results, we do not see any correlations between the Direct Bias score and the magnitude of change in the \textit{swapped} or \textit{neutralized} experiments, in other words, no strong correlation is found between the bias score and the observed effect of it in the downstream task. However, it should be noted that Direct Bias scores are generated by using pre-defined target dictionaries which might not generalize well to the downstream problem's dataset. 


\section{Discussion}
In this study, we evaluated the gender bias in 4 different pre-trained embedding alternatives and showed their implications for the downstream task with a set of experiments. We draw a few conclusions from our experiments. First, we find that depending on the dataset these embeddings are trained on, the bias direction might be different. For the target depression category, while W2VecNews embeddings are biased towards females, BioWordVec models trained on the PubMed dataset show bias towards the male group. Second, although it is not possible to directly compare non-contextual and contextual embeddings, similar to previous studies, we find that contextual embeddings have lower bias scores compared to non-contextual. Finally, we found that bias magnitude and direction in embeddings affect the false negative rates for gender groups in the downstream task. 

To analyze the effect of bias in embeddings on downstream problem, we proposed a set of experiments that changes gender information and helps us to observe the effect of different components easily for depression phenotype recognition problem. Our experiments show that when gender pronouns are removed from the training data, the model's bias direction is in line with the embeddings and we observe lower false negative rates for gender group that embeddings are biased to. 

As a bias mitigation method, we simply augmented the training dataset by gender swapping and showed that it improves the fairness of the models by increasing FNRR markedly. On the other hand, applying this method to other problems, in which text is written by the person itself and thus gender information is spread to all text implicitly, might be quite challenging (e.g. depression recognition from social media data).

\section*{Acknowledgments}
We would like to acknowledge and thank Dong Nguyen, Itir Onal Ertugrul, and Ecem Sogancioglu for their feedback which contributed to the quality of this paper.

\bibliographystyle{ACM-Reference-Format}
\bibliography{neurips}

\section{Supplementary Material}
\begin{figure}[h]
\caption{\label{fig:pca} Sorted eigenvalues (explained variance) per embedding method for the dataset obtained with the difference of embedding vectors of the gender pair words.}
     \centering
     \begin{subfigure}[b]{0.4\textwidth}
         \centering
         \includegraphics[width=\textwidth]{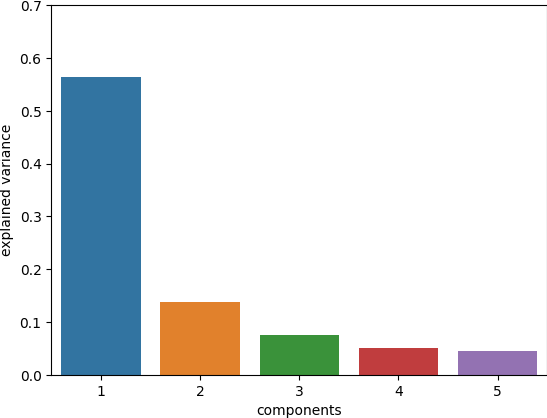}
         \caption{W-BERT}
         \label{fig:W-BERT}
     \end{subfigure}
     \hfill
     \begin{subfigure}[b]{0.4\textwidth}
         \centering
         \includegraphics[width=\textwidth]{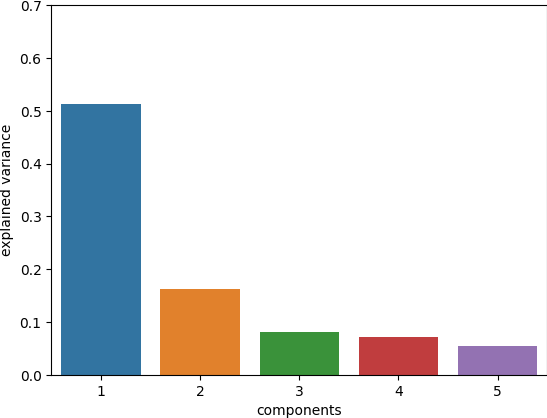}
         \caption{Clinical-BERT}
         \label{fig:Clinical-BERT}
     \end{subfigure}
     \hfill
     \begin{subfigure}[b]{0.4\textwidth}
         \centering
         \includegraphics[width=\textwidth]{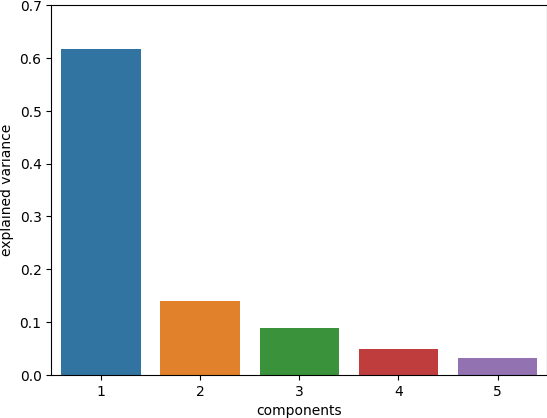}
         \caption{W2VecNews}
         \label{fig:W2VecNews}
     \end{subfigure}
     \begin{subfigure}[b]{0.4\textwidth}
         \centering
         \includegraphics[width=\textwidth]{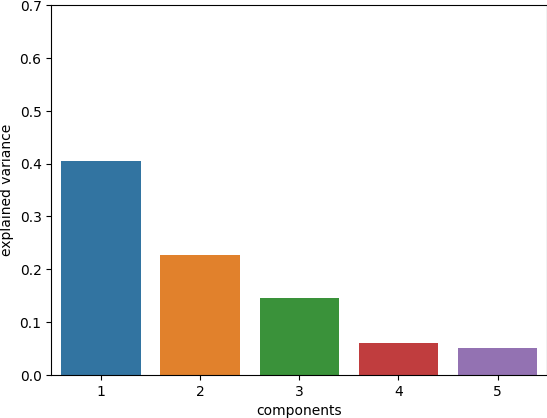}
         \caption{BioWordVec}
         \label{fig:BioWordVec}
     \end{subfigure}
        
\end{figure}

Similarly to the findings by ~\cite{bolukbasi2016man}, Fig.~\ref{fig:pca} shows that the first eigenvalue of PCA is significantly larger than the other components and that there is a single direction describing the majority of variance in all these vectors. However, we observe that the difference between the percentage of variances is higher for domain-independent embeddings compared to their clinical-domain-specific versions.

\label{sec:supp}

\begin{table*}[h]
\label{table:res}
\begin{tabular}{lllllllllllll}
\hline
\multicolumn{1}{|l|}{Model} & \multicolumn{1}{|l|}{DB} & \multicolumn{2}{l|}{original} & \multicolumn{2}{l|}{swapped} & \multicolumn{2}{l|}{neutralized} &\multicolumn{2}{l|}{augmented}\\ \hline
          & &  FNRR     & F1  & FNRR   & F1  & FNRR    & F1  & FNRR    & F1  \\

Clinical-BERT     & 0.04$_M$ & 0.93$_F$ & 0.60  & 0.91$_M$ & 0.63 & 0.92$_M$ & 0.62 & 0.94$_M$  & 0.64\\
W-BERT   & 0.02$_F$ & 0.97$_M$ & 0.60  & 0.94$_F$ & 0.58 & 1.00 & 0.58 & 0.97$_F$  &  0.58\\
BioWordVec  & 0.09$_M$ & 0.88$_M$ & 0.64 &  0.94$_M$ & 0.65 & 0.79$_M$ & 0.64 & 0.95$_F$  & 0.64\\
W2VNews    & 0.04$_F$ & 0.91$_F$ & 0.62  & 0.90$_F$ & 0.62 & 0.88$_F$ & 0.64 & 0.90$_F$ & 0.65\\ 
\hline
\end{tabular}
\caption{Performance measures of trained RF models alongside bias measures of embeddings. DB: Direct bias score of given word embedding, if embedding is biased towards the female, denoted as $_F$ otherwise $_M$, F1: macro-averaged F1 score. FNRR: false negative rate ratio. If the female groups´s FNRR is lower than the male group, this is denoted as $_F$, otherwise, $_M$.}
\end{table*}



\begin{table*}[h]
\label{table:res}
\begin{tabular}{lllllllllllll}
\hline
\multicolumn{1}{|l|}{Model} & \multicolumn{1}{|l|}{DB} & \multicolumn{2}{l|}{original} & \multicolumn{2}{l|}{swapped} & \multicolumn{2}{l|}{neutralized} &\multicolumn{2}{l|}{augmented}\\ \hline
          & &  FNRR     & F1  & FNRR   & F1  & FNRR    & F1  & FNRR    & F1  \\

Clinical-BERT     &  0.04$_M$ & 0.93$_F$ & 0.7 & 0.96$_M$ & 0.69 & 0.92$_M$ & 0.68 & 0.96$_M$ & 0.71 \\
W-BERT   & 0.02$_F$ & 0.96$_F$ &  0.66 & 0.94$_M$ & 0.66 & 0.87$_F$ & 0.64 &1.00 & 0.69 \\
BioWordVec  & 0.09$_M$ &  1.00  & 0.69  & 0.88$_F$ & 0.69 & 1.00 & 0.60 & 0.92$_F$ & 0.70\\
W2VNews   & 0.04$_F$ & 0.61$_F$  & 0.69  & 0.97$_M$ & 0.69 & 0.37$_F$ & 0.64 & 0.93$_F$  & 0.68 \\ 
\hline
\end{tabular}
\caption{Performance measures of trained MLP models alongside bias measures of embeddings. DB: Direct bias score of given word embedding, if embedding is biased towards the female, denoted as $_F$ otherwise $_M$, F1: macro-averaged F1 score. FNRR: false negative rate ratio. If the female groups´s FNRR is lower than the male group, this is denoted as $_F$, otherwise, $_M$.}
\end{table*}
\end{document}